\documentclass[11pt]{article}

\usepackage{amsmath,amssymb,amsfonts, amsbsy, amsthm, epsfig, subfig, graphicx,rotating}
\usepackage[usenames]{color}
\usepackage{enumerate}
\usepackage{hyperref}
\hypersetup{colorlinks=true, linkcolor=red, citecolor=blue}

\newcommand{\noi}{\noindent}

\newcommand{\la}{\lambda}

\newcommand{\beq}{\begin{eqnarray*}}
\newcommand{\eeq}{\end{eqnarray*}}
\newcommand{\beqn}{\begin{eqnarray}}
\newcommand{\eeqn}{\end{eqnarray}}

\newcommand{\bi}{\begin{itemize}}
\newcommand{\ei}{\end{itemize}}

\newcommand{\be}{\begin{equation}}
\newcommand{\ee}{\end{equation}}
\newcommand{\nn}{\nonumber}

\newcommand{\lbl}{\label}

\newcommand{\bbox}{\nobreak\quad\vrule width4pt depth2pt height4pt}
\newcommand{\eq}[1]{$(\ref{#1})$}

\newcommand{\ignore}[1]{}{}
\newcommand{\f}{\frac}

\newcommand{\sn}{\sum_{i=1}^n}

\renewcommand{\P}{\mathbb{P}}

\newcommand{\de}{\delta}
\newcommand{\De}{\Delta}
\newcommand{\ga}{\gamma}

\newcommand{\s}{\sqrt}

\newcommand{\e}{\mathbb{E}}
\newcommand{\p}{\mathbb{P}}

\newcommand{\br}{\mathbb{R}}

\newcommand{\argmin}{\mathop{\rm arg\min}}

\numberwithin{equation}{section}
\theoremstyle{plain}

\newtheorem{lemma}{Lemma}[section]
\newtheorem{proposition}{Proposition}[section]
\newtheorem{theorem}{Theorem}[section]

\newtheorem{remark}{Remark}[section]
\newtheorem{definition}{Definition}[section]

\begin{document}

\title{A Max-Norm Constrained Minimization Approach to 1-Bit Matrix Completion}
\author{T. Tony Cai$^{1}$ and Wen-Xin Zhou$^{2,3}$}
\date{}
\maketitle

\begin{abstract}
We consider in this paper the problem of noisy 1-bit matrix completion under a general non-uniform sampling distribution using the max-norm as a convex relaxation for the rank. A max-norm constrained maximum likelihood estimate is introduced and studied. The rate of convergence for the estimate  is obtained. Information-theoretical methods are used to establish a minimax lower bound  under the general sampling model. The minimax upper and lower bounds together yield the optimal rate of convergence for the Frobenius norm loss. Computational algorithms  and numerical performance are also discussed.
\end{abstract}

\footnotetext[1]{
Department of Statistics, The Wharton School, University of Pennsylvania, Philadelphia, PA 19104.  \newline
\indent \ \
 The research of Tony Cai was supported in part by NSF FRG Grant DMS-0854973, NSF Grant DMS \newline \indent \  \ -1208982,  and NIH Grant R01 CA 127334-05.}
\footnotetext[2]{
Department of Mathematics, Hong Kong University of Science and Technology, Clear Water Bay, \newline \indent \ \
 Kowloon, Hong Kong.}
\footnotetext[3]{
Department of Mathematics and Statistics, University of Melbourne, Parkville, VIC, 3010, Australia.}

\noindent{\bf Keywords:\/}  1-bit matrix completion, Frobenius norm,  low-rank matrix,  max-norm,  constrained optimization, maximum likelihood estimate, optimal rate of convergence, trace norm.

\newpage

\section{Introduction}
\setcounter{equation}{0}

Matrix completion, which aims to recover a low-rank matrix from a subset of its entries, has been an active area of research in the last few years. It has a range of successful applications. In some real-life situations, however, the observations are highly \textit{quantized}, sometimes even to a single bit and thus the standard matrix completion techniques do not apply. Take the Netflix problem as an example, the observations are the ratings of movies, which are quantized to the set of integers from $1$ to $5$. In the more extreme case such as recommender systems, only a single bit of rating standing for a ``thumbs up'' or ``thumbs down'' is recorded at each occurrence. Another example of applications is targeted advertising, such as the relevance of advertisements on Hulu. Each user who is watching TV shows on Hulu is required to answer yes/no to the question{\it ``Is this ad relevant to you?''}. Noise effect should be considered since there  are users who just click no to all the advertisements. In general, people would prefer to have advertisement catered to them, rather than to endure random advertisement.  Targeted marketing that utilizes customer needs tends to serve better than random, scattershot advertisements. Similar idea has already been employed in mail system \cite{GNOT92}. Other examples from recommender systems include rating music on Pandora and posts on Reddit or MathOverflow, in which each observation consists of a single bit representing a positive or negative rating. Similar problem also arises in analyzing incomplete survey designs containing simple agree/disagree questions in the analysis of survey data, and distance matrix recovery in multidimensional scaling using binary and incomplete data \cite{GW73, SD74}. See \cite{DP} for more detailed discussions.

Motivated by these applications, Davenport, {\it et al.} (2012) considered the {\it 1-bit matrix completion problem} of recovering an approximately low-rank matrix $M^* \in \br^{d_1 \times d_2}$ from a set of $n$ noise corrupted sign (i.e., 1-bit) measurements. In particular, they proposed a trace-norm constrained maximum likelihood estimator to estimate $M^*$, based on a small number of binary samples observed according to a probability distribution determined by the entries of $M^*$. It was also shown that the trace-norm constrained optimization method is minimax rate-optimal under the uniform sampling model. This problem is closely connected to and in some respects more challenging than the {\it 1-bit compressed sensing}, which was introduced and first studied by Boufounos and Baraniuk (2008). The 1-bit measurements are meant to model quantization in the extreme case, and a surprising fact is that when the signal-to-noise ratio is low, empirical evidence demonstrates that such extreme quantization can be optimal when constrained to a fixed bit budget \cite{LB}. We refer to \cite{Plan12} for a list of growing literature on 1-bit compressed sensing.

To be more specific, consider an arbitrary unknown $d_1 \times d_2$ target matrix $M^*$ with rank at most $r$. Suppose a subset $S=\{(i_1, j_1), ..., (i_n, j_n)\}$ of entries of a binary matrix $Y$ is observed, where the entries of $Y$ depend on $M^*$ in the following way:
\beqn
	Y_{i,j}  = \left\{\begin{array}{ll}
	+1,    & \mbox{if } M^*_{i , j}+Z_{i , j} \geq 0,  \\
	-1,   &  \mbox{if } M^*_{i , j}+Z_{i , j} < 0.
	\end{array}  \right.  \lbl{model0}
\eeqn
Here $Z=(Z_{ij})\in \br^{d_1 \times d_2}$ is a general noise matrix. This latent variable matrix model can been seen as a direct analogue to the usual 1-bit compressed sensing model, in which only the signs of measurements are observed. It is known that an $s$-sparse signal can still be approximately recovered from $O(s \log(d/s))$ random linear measurements. See, e.g. \cite{Jac11, Plan11, Plan12,  Ai12}.

Contrary to the standard matrix completion model and many other statistical problems,  random noise turns out to be helpful and has a positive effect in the 1-bit case, since the problem is ill-posed in the absence of noise as described in \cite{DP}. In particular, when $Z=0$ and $M^* = u v^T$ for some vectors $u \in \br^{d_1}, v\in \br^{d_2}$ having no zero coordinates, then the radically disparate matrix $\tilde{M} = \mbox{sign}(u) \mbox{sign}^T(v)$ will lead to the same observations $Y$. Thus $M$ and $\tilde{M}$ are indistinguishable. However, it has been surprisingly noticed that the problem may become well-posed when there are some additional stochastic variations, that is, $Z \neq 0$ is an appropriate random noise matrix. This phenomenon can be regarded as a ``dithering'' effect brought by random noise.

Although the trace-norm constrained optimization method has been shown to be minimax rate-optimal under the uniform sampling model, it remains unclear that the trace-norm is the best convex surrogate to the rank. A different convex relaxation for the rank, the matrix {\it max-norm}, has been duly noted in machine learning literature since Srebro, Rennie and Jaakkola (2004), and it was shown to be empirically superior to the trace-norm for collaborative filtering problems. Regarding a real $d_1 \times d_2$ matrix as an operator that maps from $\br^{d_2}$ to $\br^{d_1}$, its rank can be alternatively expressed as the smallest integer $k$, such that it is possible to express $M=U V^T$, where $U \in \br^{d_1\times k}$ and $V \in \br^{d_2 \times k}$. In terms of the matrix factorization $M=UV^T$, we would like $U$ and $V$ to have a small number of columns. The number of columns of $U$ and $V$ can be relaxed in a different way from the usual trace-norm by the so-called {\it max-norm} \cite{LMSS04} which is defined by
\be
	\| M \|_{\max} = \min_{M=UV^T}  \big\{ \| U \|_{2, \infty} \| V \|_{2,\infty}  \big\},
	 \lbl{maxnorm}
\ee
where the infimum is over all factorizations $M=UV^T$ with $\| U \|_{2, \infty}$ being the operator norm of $U: \ell_2^k \rightarrow \ell^{d_1}_{\infty}$ and $ \| V \|_{2,\infty} $ the operator norm of $V:\ell_2^k \rightarrow \ell_{\infty}^{d_2}$ (or, equivalently, $V^T: \ell^{d_2}_1 \rightarrow \ell_2^k$) and $k=1, ..., \min(d_1, d_2)$. It is not hard to check that $\|U \|_{2,\infty}$ is equal to the largest $\ell_2$ norm of the rows in $U$. Since $\ell_2$ is a Hilbert space, $\| \cdot \|_{\max}$ indeed defines a norm on the space of operators between $\ell^{d_2}_1$ and $\ell_{\infty}^{d_1}$. Comparably, the trace-norm has a formulation similar to \eq{maxnorm}, as given below in Section 2.1.

Foygel and Srebro (2011)  first used the max-norm for matrix completion under the uniform sampling distribution. Their results are direct consequences of a recent bound on the excess risk for a smooth loss function, such as the quadratic loss, with a bounded second derivative \cite{SST10}. Matrix completion under a non-degenerate random sampling model was considered by the present authors in an earlier paper \cite{CaiZ12}. It was shown that the max-norm constrained minimization method is rate-optimal and it yields a more stable approximate recovery guarantee, with respect to the sampling distributions, than trace-norm based approaches.

Davenport, \textit{et al.} (2012)  analyzed 1-bit matrix completion under the {\it uniform sampling model}, where observed entries are assumed to be sampled randomly and uniformly. In such a setting, the trace-norm constrained approach has been shown to achieve minimax rate of convergence. However, in certain application such as collaborative filtering, the uniform sampling model is over idealized. In the Netflix problem, for instance, the uniform sampling model is equivalent to assuming all users are equally likely to rate every movie and all movies are equally likely to be rated by any user. In practice, inevitably some users are more active than others and some movies are more popular and thus rated more frequently. Therefore, the sampling distribution is in fact non-uniform. In such a scenario, Salakhutdinov and Srebro (2010)  showed that the standard trace-norm relaxation can behave very poorly, and suggested to use a weighted variant of the trace-norm, which takes the sampling distribution into account. Since the true sampling distribution is most likely unknown and can only be estimated based on the locations of those entries that are revealed in the sample, what commonly used in practice is the empirically-weighted trace norm. Foygel, {\it et al.} (2011)  provided rigorous recovery guarantees for learning with the standard weighted, smoothed weighted and smoothed empirically-weighted trace-norms. In particular, they gave upper bounds on excess error, which show that there is no theoretical disadvantage of learning with smoothed empirical marginals as compared to learning with smoothed true marginals.

In this paper we study matrix completion based on noisy 1-bit observations under a general (non-degenerate) sampling model using the max-norm as a convex relaxation for the rank. The rate of convergence for the max-norm constrained maximum likelihood estimate is obtained. A matching minimax lower bound  is established under the general non-uniform sampling model using information-theoretical methods. The minimax upper and lower bounds together yield the optimal rate of convergence for the Frobenius norm loss. As a comparison with the max-norm constrained optimization approach, we also analyze the recovery guarantee of the weighted trace-norm constrained method in the setting of non-uniform sampling distributions. Our result includes an additional logarithmic factor, which might be an artifact of the proof technique. To sum up, the max-norm regularized approach indeed provides a unified and stable approximate recovery guarantee with respect to the sampling distributions, while previously used approaches are based on different variants of the trace-norm which may sometimes seem artificial to practitioners.

When the noise distribution is Gaussian or more generally log-concave, the negative log-likelihood function for $M$, given the measurements, is convex, hence computing the max-norm constrained maximum likelihood estimate is a convex optimization problem. The computational effectiveness of this method is also studied, based on a first-order algorithm developed in \cite{Lee2010} for solving convex programs involving a max-norm constraint, which outperforms the semi-definite programming method of Srebro, {\it et al.} (2004). It will be shown in Section \ref{implementation.sec} that the convex optimization problem can be implemented in polynomial time as a function of the sample size and the matrix dimensions.

The rest of the paper is organized as follows. Section \ref{notation.sec} begins with the basic notation and definitions, and then states a collection of useful results on the matrix norms, Rademacher complexity and distances between matrices that will be needed throughout the paper. Section \ref{main.sec} introduces the 1-bit matrix completion model and the estimation procedure and investigates the theoretical properties of the estimator. Both minimax upper and lower bounds are established. The results show that the max-norm constraint maximum likelihood estimator is rate-optimal over the parameter space. Section \ref{main.sec} also gives a comparison of our results with previous work. Computational algorithms  are discussed in Section \ref{implementation.sec}, and numerical performance of the proposed algorithm is considered in Section \ref{sim.sec}. The proofs of the main results are given in Section \ref{proof.sec}. The paper is concluded with a brief discussion in Section \ref{dscs.sec}.

\section{Notations and Preliminaries}
\label{notation.sec}
\setcounter{equation}{0}

In this section, we introduce basic notation and definitions that will be used throughout the
paper, and state some known results on the max-norm, trace-norm and Rademacher complexity that will be used repeatedly later.

\medskip

\noi
{\sc Notation.} For any positive integer $d$, we use $[d]$ to denote the set of integers $\{ 1, 2, ..., d\}$. For any pair of real numbers $a$ and $b$, set $a \vee b := \max(a, b)$ and $a \wedge b := \min(a, b)$. For a vector $u\in \br^d$ and $0<p < \infty$, denote its $\ell_p$-norm by $\| u \|_p=(\sum_{i=1}^d |u_i|^p)^{1/p}$. In particular, $\| u \|_{\infty}=\max_{i=1, ..., d}| u_i|$ is the $\ell_{\infty}$-norm. For a matrix $M =(M_{k,l}) \in \br^{d_1 \times d_2}$, let $\| M \|_F = \s{\sum_{k=1}^{d_1} \sum_{l=1}^{d_2} M_{k,l}^2}$ be the Frobenius norm and let $\| M\|_{\infty}=\max_{k,l}|M_{k,l}|$ denote the elementwise $\ell_{\infty}$-norm. Given two norms $\ell_p$ and $\ell_q$ on $\br^{d_1}$ and $\br^{d_2}$ respectively, the corresponding operator norm $\| \cdot \|_{p,q}$ of a matrix $M\in \br^{d_1 \times d_2}$  is defined by $\| M \|_{p,q}=\sup_{\| x\|_p=1} \| M x\|_q$.  It is easy to verify that $\| M\|_{p,q}=\| M^T \|_{q^*,p^*}$, where $(p,p^*)$ and $(q, q^*)$ are conjugate pairs, i.e. $\f{1}{p}+\f{1}{p^*}=1$ and $\f{1}{q}+\f{1}{q^*}=1$. In particular, $\| M \| = \| M \|_{2,2}$ is the spectral norm and $\| M \|_{2,\infty}=\max_{k=1, ..., d_1} \s{\sum_{l=1}^{d_2} M_{k,l}^2}$ is the maximum row norm of $M$.

\medskip
\subsection{Max-norm and trace-norm} \lbl{pre-1}

For any matrix $M \in \br^{d_1 \times d_2}$, its \textit{trace-norm} is defined to be the sum of the singular values of $M$ (i.e. the roots of the eigenvalues of $M M^T$), and can also equivalently written as
\beq
	\| M \|_* = \inf \bigg\{ \sum_j |\sigma_j | :  M = \sum_j \sigma_j u_j v_j^T, \,  u_j \in \br^{d_1}, v_j \in \br^{d_2} \mbox{ satisfying } \| u_j \|_2 = \| v_j \|_2 =1 \bigg\}.
\eeq
Recall the definition \eq{maxnorm} of the max-norm, the trace-norm can be analogously defined in terms of matrix factorization as
$$
	\| M \|_* = \min_{M = U V^T} \, \big\{ \| U \|_F \| V \|_F \big\} = \f{1}{2} \min_{U, V: M = U V^T} \big( \| U \|_F^2 + \| V \|_F^2 \big).
$$
Since the $\ell_1$-norm of a vector is bounded by the product of its $\ell_2$-norm and the number of non-zero coordinates, we have the following relationship between the trace-norm and Frobenius norm
\beq
	\|M \|_F \leq \| M \|_* \leq \s{\mbox{rank}(M)} \cdot \| M \|_F.
\eeq

By the elementary inequality $\| M_{m\times n} \|_F \leq \s{m} \| M_{m \times n} \|_{2, \infty}$, we see that
\be
	\f{\| M \|_*}{\s{d_1 d_2}}  \leq \| M \|_{\max}.  \lbl{tr-max}
\ee
Furthermore, as was noticed in Lee, \textit{et al.} (2010), the max-norm, which is defined in \eq{maxnorm}, is comparable with a trace-norm more precisely in the following sense \cite{J87}:
\beqn
	\lefteqn{	\| M \|_{\max}    }  \lbl{mn.decom} \\
	&& \approx  \inf\bigg\{ \sum_j |\sigma_j | :  M = \sum_j \sigma_j u_j v_j^T, \,  u_j \in \br^{d_1}, v_j \in \br^{d_2} \mbox{ satisfying } \| u_j \|_{\infty} = \| v_j \|_{\infty} =1 \bigg\},   \nn
\eeqn
where the factor of equivalence is $K_G \in (1.67, 1.79)$, denoting the Grothendieck's constant. What may be more surprising is the following bounds for the max-norm, in connection with element-wise $\ell_{\infty}$-norm \cite{LMSS04}:
\be
	\| M \|_{\infty } \leq \| M \|_{\max} \leq \s{\mbox{rank}(M)} \cdot \| M \|_{1, \infty} \leq \s{\mbox{rank}(M)} \cdot \| M \|_{\infty}.  \lbl{m-i}
\ee

\medskip
\subsection{Rademacher complexity}

Considering matrices as functions from index pairs to entry values, a technical tool used in our proof involves data-dependent estimates of the {\it Rademacher complexity} of the classes that consist of low trace-norm and low max-norm matrices. We refer to Bartlett and Mendelson (2002) for a detailed introduction of this concept.

\begin{definition}
Let $\mathcal{P}$ be a probability distribution on a set $\mathcal{X}$. Suppose that $X_1, ..., X_n$ are independent samples drawn from $\mathcal{X}$ according to $\mathcal{P}$, and set $S=\{ X_1, ..., X_n\}$. For a class $\mathcal{F}$ of functions mapping from $\mathcal{X}$ to $\mathbb{R}$, its {\it empirical Rademacher complexity} over the sample $S$ is defined by
\be
	\hat{R}_{S}(\mathcal{F}) = \f{2}{|S|}  \e_{\varepsilon} \Big[ \sup_{f\in \mathcal{F}} \Big| \sn \varepsilon_i f(X_i) \Big| \Big],  \lbl{rc}
\ee
where $\varepsilon=(\varepsilon_1, ..., \varepsilon_n)$ is a Rademacher sequence. The Rademacher complexity with respect to the distribution $\mathcal{P}$ is the expectation, over a sample $S$ of $|S|$ points drawn i.i.d. according to $\mathcal{P}$, denoted by
$$
	R_{|S|} (\mathcal{F}) =	\e_{S \sim \mathcal{P}}[\hat{R}_{S}(\mathcal{F})].
$$
\end{definition}

The following properties regarding $\hat{R}_{S}(\mathcal{F})$ are useful.

\begin{proposition}
We have
\begin{enumerate}
\item[1.] If $\mathcal{F} \subseteq \mathcal{G}$, $\hat{R}_{S}(\mathcal{F}) \leq \hat{R}_{S}(\mathcal{G})$.

\item[2.] $\hat{R}_{S}(\mathcal{F}) = \hat{R}_{S}(\mbox{\rm conv} (\mathcal{F}))=\hat{R}_{S}(\mbox{\rm absconv} (\mathcal{F}))$, where $\mbox{\rm conv} (\mathcal{F})$ is the class of convex combinations of functions from $\mathcal{F}$, and $\mbox{\rm absconv} (\mathcal{F})$ denotes the absolutely convex hull of $\mathcal{F}$, that is, the class of convex combinations of functions from $\mathcal{F}$ and $-\mathcal{F}$.

\item[3.] For every $c \in \br$, $\hat{R}_{S}( c \mathcal{F}) = |c| \hat{R}_{S}(\mathcal{F})$, where $c \mathcal{F} \equiv \{c f: f \in \mathcal{F}\}$.
\end{enumerate}
\end{proposition}

In particular, we are interested in calculating the Rademacher complexities of the trace-norm and max-norm balls. To this end, define for any radius $R>0$ that
\beq
	\mathbb{B}_*(R)  & := & \big\{ M \in \br^{d_1 \times d_2}: \| M \|_* \leq R \big\}
	\ \  \mbox{ and }  \\
	\mathbb{B}_{\max}(R)  &:= & \big\{ M \in \br^{d_1 \times d_2}: \| M \|_{\max} \leq R \big\}.
\eeq
First, recall that any matrix with unit trace-norm is a convex combination of unit-norm rank-one matrices, and thus
\be
	\mathbb{B}_*(1) = \mbox{conv} (\mathcal{M}_1), \ \ \mbox{ where }
	\mathcal{M}_1 :=  \big\{ u v^T:  u \in \br^{d_1}, v \in \br^{d_2}, \| u \|_2 = \| v \|_2 =1 \big\}.
	\lbl{tn.cvhull}
\ee
Then $\hat{R}_{S}(\mathbb{B}_*(1)) = \hat{R}_{S}(\mathcal{M}_1)$. A sharp bound on the worst-case Rademacher complexity, defined as the supremum of $\hat{R}_{S}(\cdot )$ over all sample sets $S$ with size $|S|=n$, is $\f{2}{\s{n}}$ (See, expression (4) on Page 551, \cite{SS2005}). This bound, unfortunately, is barely useful in developing generalization error bounds. However, when the index pairs of a sample $S$ are drawn uniformly at random from $[d_1]\times [d_2]$ (with replacement), Srebro and Shraibman (2005) showed that the {\it expected} Rademacher complexity is low, and Foygel and Srebro (2010) have improved this result by reducing the logarithmic factor. In particular, they proved that for a sample size $n \geq d=d_1+d_2$,
\be
	\e_{S \sim \mbox{unif}, |S|=n} \big[ \hat{R}_S ( \mathbb{B}_*(1) ) \big] \leq \f{K}{\s{d_1 d_2}} \s{\f{d \log(d)}{ n}},	\lbl{rc-bd}
\ee
where $K>0$ denotes a universal constant.

The unit max-norm ball, on the other hand, can be approximately characterized as a convex hull. Due to the Grothendieck's inequality, it was shown in \cite{SS2005} that
\be
	 \mbox{conv} (\mathcal{M}_{\pm}) \subset \mathbb{B}_{\max}(1)  \subset  K_G \cdot \mbox{conv} (\mathcal{M}_{\pm}),  \lbl{mn.cvhull}
\ee
where $\mathcal{M}_{\pm} := \{ M \in \{\pm 1 \}^{d_1 \times d_2}: \mbox{rank}(M)=1\}$ is the class of rank-one sign matrices, and $ K_G \in (1.67, 1.79)$ is the Grothendieck's constant. It is easy to see that $\mathcal{M}_{\pm}$ is a finite class with cardinality $|\mathcal{M}_{\pm}|=2^{d-1}$, $d=d_1+d_2$. For any $d_1, d_2>2$ and any sample of size $2<|S| \leq d_1 d_2$, the empirical Rademacher complexity of the unit max-norm ball is bounded by
\be
	\hat{R}_S \big( \mathbb{B}_{\max}(1) \big) \leq 12  \s{ \f{d}{|S|} }. \lbl{erc-bd}
\ee
In other words, $\sup_{S: |S|=n}\hat{R}_S ( \mathbb{B}_{\max}(1) ) \leq 12  \s{ \f{d}{n} }$.

\medskip
\subsection{Discrepancy}

In order to get both upper and lower prediction error bounds on the weighted squared Frobenius norm between the proposed estimator, given by \eq{max-est} below,  and the target matrix described via model \eq{1b}, we will need the following two concepts of discrepancies between matrices as well as their connections. In particular, we will focus on element-wise notion of discrepancy between two $d_1 \times d_2$ matrices $P$ and $Q$.

First, for two matrices $P$, $Q : [d_1]\times [d_2] \rightarrow [0,1]^{d_1 \times d_2}$, their Hellinger distance is given by
\beq
	d_H^2(P; Q) = \f{1}{d_1 d_2} \sum_{(k,l)} d_H^2(P_{k,l};Q_{k,l}),
\eeq
where $d_H^2(p;q)=(\s{p} - \s{q})^2 + (\s{1-p} - \s{1-q})^2$ for $p, q \in [0,1]$. Next, the Kullback-Leibler divergence between two matrices $P$, $Q : [d_1]\times [d_2]  \rightarrow [0,1]^{d_1 \times d_2}$ is defined by
\beq
	\mathbb{K}(P \| Q) = \f{1}{d_1 d_2} \sum_{(k, l)} K(P_{k,l} \| Q_{k,l}),
\eeq 	
where $K(p \| q ) = p\log(\f{p}{q})+(1-p)\log({\f{1-p}{1-q}})$, for $p, q \in [0,1]$. Note that $\mathbb{K}(P \| Q)$ is not a distance; it is sufficient to observe that it is not symmetric.

The relationship between the two ``distances'' is as follows. For any two scalars $p, q\in [0,1]$, we have
\be
	d_H^2(p; q) \leq K(p \| q),  \lbl{dis0}
\ee
which in turn implies that, for any two matrices $P$, $Q : [d_1]\times [d_2] \rightarrow [0,1]^{d_1\times d_2}$,
\be
		d_H^2(P; Q)  \leq  \mathbb{K}(P \| Q).  \lbl{dis}
\ee
The proof of \eq{dis0} is based on the Jensen's inequality and an elementary inequality that $1-x \leq -\log x$ for any $x>0$.

\section{Max-Norm Constrained Maximum Likelihood Estimate}
\setcounter{equation}{0}
\label{main.sec}

In this section, we introduce the max-norm constrained maximum likelihood estimation procedure for 1-bit matrix completion and investigates the theoretical properties of the estimator. The results are also compared with other results in the literature.

\subsection{Observation model}

We consider 1-bit matrix completion under a general random sampling model. The unknown low-rank matrix $M^* \in \br^{d_1 \times d_2}$ is the object of interest. Instead of observing noisy entries $M^*_{i,j}+Z_{i,j}$ directly in {\it unquantized} matrix completion, now we only observe with error the sign of a random subset of the entries of $M^*$. More specifically, assume that a random sample
$$
S= \big\{ (i_1, j_1), (i_2, j_2), ..., (i_n,j_n)  \big\}  \subseteq \big(  [d_1] \times [d_2] \big)^n
$$
of the index set is drawn  i.i.d. with replacement according to a general sampling distribution $\Pi=\{\pi_{kl}\}$ on $[d_1]\times [d_2]$. That is, $\p\{(i_t, j_t)=(k,l)\} = \pi_{kl}$, for all $t$ and $(k, l)$. Suppose that a (random) subset $S$ of size $|S|=n$ of entries of a sign matrix $Y$ is observed. The dependence of $Y$ on the underlying matrix $M^*$ is as follows:
\beqn
	Y_{i, j}  = \left\{\begin{array}{ll}
		+1,    & \mbox{if } M^*_{i, j}+ Z_{i,j} \geq 0,  \\
		-1, &  \mbox{if } M^*_{i,j}+ Z_{i,j} < 0,
		\end{array}  \right.    \lbl{1b}
\eeqn
where $Z=(Z_{i,j}) \in \br^{d_1 \times d_2}$ is a matrix consisting of i.i.d. noise variables. Let $F(\cdot )$ be the cumulative distribution function of $-Z_{1,1}$, then the above model can be recast as
\beqn
	Y_{i, j}  = \left\{
								\begin{array}{ll}
											+1,    & \mbox{ with probability } F(M^*_{i, j}),  \\
											-1,
											  &  \mbox{ with probability } 1- F(M^*_{i, j}),
									\end{array}  \right.  \ \     \lbl{1b-md}
\eeqn
and we observe noisy entries $\{ Y_{i_t, j_t} \}_{t=1}^n$ indexed by $S$. More generally, we consider the model \eq{1b-md} with an arbitrary differentiable function $F: \br \rightarrow [0,1]$. Particular assumptions on $F$ will be discussed below.

Instead of assuming the uniform sampling distribution \cite{DP}, here we allow a general sampling distribution $\Pi = \{\pi_{kl}\}$, satisfying $\sum_{(k,l) \in [d_1] \times [d_2]} \pi_{kl}=1$, according to which we make $n$ independent random choices of entries. The drawback of the setting is that, with fairly high probability, some entries will be sampled multiple times. Intuitively it would be more practical to assume that entries are sampled without replacement, or equivalently, to sample $n$ of the $d_1 d_2$ binary entries observed with noise without replacing. Due to the requirement that the drawn entries be distinct, the $n$ samples are not independent. This dependence structure turns out to impede the technical analysis of the learning guarantees. To avoid this complication, we will use the i.i.d. approach as a proxy for sampling without replacement throughout this paper. As has been noted in \cite{GN10,FS2011}, between sampling with and without replacement both in a uniform sense, that is, making $n$ independent uniform choices of entries versus choosing a set $S$ of entries uniformly at random over all subsets that consist of exactly $n$ entries, the latter is indeed as good as the former. See Sect.~\ref{Note} below for more details.

Next we list three natural choices for $F$, or equivalently, for the distribution of $\{Z_{i,j}\}$.

\medskip
\noi
{\sc Examples}:

\begin{enumerate}
\item[1.](Logistic regression/Logistic noise): The logistic regression model is described by \eq{1b-md} with
$$F(x)=\f{e^x}{1+e^x}, $$ and equivalently by \eq{1b} with $Z_{i,j}$ i.i.d. following the standard logistic distribution.

\item[2.](Probit regression/Gaussian noise): The probit regression model is described by \eq{1b-md} with
$$ F(x)=\Phi\Big(\f{x}{\sigma} \Big), $$ where $\Phi$ denotes the cumulative distribution function of $N(0,1)$, and equivalently by \eq{1b} with $Z_{i,j}$ i.i.d. following $N(0, \sigma^2)$.

\item[3.](Laplace noise): Another interesting case is that $Z_{i,j}$'s are i.i.d. Laplace noise (Laplace$(0,b)$), with
\beq
	F(x)  = \left\{
								\begin{array}{ll}
											\f{1}{2}\exp(x/b),    & \mbox{ if } x<0,  \\
											1-\f{1}{2}\exp(-x/b),	  &  \mbox{ if } x \geq 0,
									\end{array}  \right.  \ \
\eeq
where $b>0$ is the scale parameter.
\end{enumerate}

Davenport, {\it et al.} (2012) have focused on approximately low-rank matrices recovery by considering the following class of matrices
\be
		K_*( \alpha , r ) = \Big\{ M \in \br^{d_1 \times d_2}: \| M \|_{\infty} \leq \alpha, \; \f{\| M \|_*}{\s{d_1 d_2}} \leq \alpha \s{r}  \Big\},  \lbl{tn.space}
\ee
where $1\leq r\leq \min(d_1, d_2)$ and $\alpha>0$ is a free parameter to be determined. Clearly, any matrix $M$ with rank at most $r$ satisfying $\| M \|_{\infty} \leq \alpha$ belongs to $K_*( \alpha , r )$. Alternatively, using max-norm as a convex relaxation for the rank, we consider recovery of matrices with $\ell_{\infty}$-norm and max-norm constraints defined by
\be
	K_{\max}(\alpha , R) := \Big\{ M\in \br^{d_1 \times d_2} : \| M \|_{\infty} \leq \alpha, \; \| M \|_{\max} \leq R \Big\}. \lbl{mn.space}
\ee
Here both $\alpha>0$ and $R>0$ are free parameters to be determined. If $M^*$ is of  rank at most $r$ and $\| M^* \|_{\infty} \leq \alpha$, then by \eq{tr-max} and \eq{m-i} we have $M^* \in \mathbb{B}_{\max}(\alpha\s{r})$ and hence
$$
	M^* \in K_{\max}(\alpha, \alpha \s{r} )  \subset K_*(\alpha, r).
$$

\subsection{Max-norm constrained maximum likelihood estimate}

Now, given a collection of observations $Y_S = \{Y_{i_t, j_t} \}_{t=1}^n$ from the observation model \eq{1b-md}, the negative log-likelihood function can be written as
\beq
	\ell_S(M;Y) = \sum_{t=1}^n \bigg[ \textbf{1}_{\{Y_{i_t ,j_t}=1\}}\log\Big( \f{1}{ F(M_{i_t, j_t})}\Big)  + \textbf{1}_{\{Y_{i_t, j_t}=-1\}}\log \Big( \f{1}{1-F(M_{i_t, j_t}) }\Big) \bigg].
\eeq
Then we consider estimating the unknown $M^* \in K_{\max}(\alpha, R)$ by maximizing the empirical likelihood function subject to a max-norm constraint, i.e.,
\be
	\hat{M}_{\max} =  \argmin_{M \in K_{\max}(\alpha , R)} \, \ell_S(M;Y).  \lbl{max-est}
\ee
The optimization procedure requires that all the entries of $M_0$ are bounded in absolute value by a  pre-defined constant $\alpha$. This condition is reasonable while also critical in approximate low-rank matrix recovery problems by controlling the {\it spikiness} of the solution. Indeed, the measure of the ``spikiness'' of matrices is much less restrictive than the incoherence conditions imposed in exact low-rank matrix recovery. See, e.g. \cite{KLT2011, NW2012, Klo12, CaiZ12}.

As has been noted before (Srebro and Shraibman, 2005), a large gap between the max-complexity (related to max-norm) and the dimensional-complexity (related to rank) is possible only when the underlying low-rank matrix has entries of vastly varying magnitudes. Also, in view of \eq{mn.decom}, the max-norm promotes low-rank decomposition with factors in $\ell_{\infty}$ ($\ell_2$ for the trace-norm). Motivated by these features, max-norm regularization is expected to be reasonably effective for uniformly bounded data.

When the noise distribution is log-concave so that the log-likelihood is a concave function, the max-norm constrained minimization problem \eq{max-est} is a convex program and we recommend a fast and efficient algorithm developed in \cite{Lee2010} for solving large-scale optimization problems that incorporate the max-norm. We will show in Section \ref{implementation.sec} that the convex optimization problem \eq{max-est} can indeed be implemented in  polynomial time as a function of the sample size $n$ and the matrix dimensions $d_1$ and $d_2$.

\subsection{Upper bounds}

To establish an upper bound on the prediction error of estimator $\hat{M}_{\max}$ given by \eq{max-est}, we need the following assumption on the unknown matrix $M^*$ as well as the regularity conditions on the function $F$ in \eq{1b-md}. \\

\noi
{\bf  Condition U}: Assume that there exist positive constants $R$ and $\alpha$ such that
\begin{enumerate}
\item[(U1)] $M^* \in K_{\max}(\alpha, R )$;
\item[(U2)] $F$ and $F'$ are non-zero in $[-\alpha, \alpha]$, and
\item[(U3)] both
\be
	L_{\alpha} := \sup_{|x|\leq \alpha} \f{|F'(x)|}{F(x)(1-F(x))}, \ \ \mbox{ and } \ \
	\beta_{\alpha} := \sup_{|x| \leq \alpha} \f{F(x)(1-F(x))}{(F'(x))^2} \lbl{reg}
\ee
are finite.
\end{enumerate}

In particular under condition $(U2)$, the quantity
\be
	U_{\alpha}  := \sup_{|x|\leq \alpha}  \log\bigg(\f{1}{F(x)(1-F(x))} \bigg),
	 \lbl{reg2}
\ee
is well-defined. As prototypical examples, we specify below the quantities $L_{\alpha}$, $\beta_{\alpha}$ and $U_{\alpha}$ in the cases of Logistic, Gaussian and Laplace noise:
\begin{enumerate}
\item[1.](Logistic regression/Logistic noise): For $F(x)=e^x/(1+e^x)$, we have
\be
L_{\alpha} \equiv 1, \ \  \beta_{\alpha}=\f{(1+e^{\alpha})^2}{e^{\alpha}} \ \ \mbox{ and } \ \
U_{\alpha}  =  2\log(e^{\alpha/2}+e^{-\alpha/2}).   \lbl{logis-ex}
\ee

\item[2.](Probit regression/Gaussian noise): For $F(x)=\Phi(x/\sigma)$, straightforward calculations show that
\be
L_{\alpha} \leq \f{4}{\sigma} \Big( \f{\alpha}{\sigma} + 1 \Big) , \ \
\beta_{\alpha}  \leq \pi \sigma^2 \exp\{\alpha^2/(2\sigma^2)\} \ \    \mbox{ and }  \  \
U_{\alpha}  \leq  \Big( \f{\alpha}{\sigma} +1 \Big)^2.   \lbl{prob-ex}
\ee

\item[3.](Laplace noise): For a Laplace$(0,b)$ distribution function, we have
\be
L_{\alpha} = \f{2}{b} , \ \
\beta_{\alpha}  = b\big( 2\exp(\alpha/b) -1 \big) \ \    \mbox{ and }  \  \
U_{\alpha}  \leq  2\Big( \f{\alpha}{ b} +\log 2 \Big).
\ee
\end{enumerate}

Now we are ready to state our main results concerning the recovery of an approximately low-rank matrix $M^*$ using the max-norm constrained maximum likelihood estimate. We write hereafter $d = d_1 + d_2$ for brevity.

\begin{theorem} \lbl{thm1-1b}
Suppose that {\rm Condition U} holds and assume that the training set $S$ follows a general weighted sampling model according to the distribution $\Pi$. Then there exists an absolute constant $C$ such that, for a sample size $2< n \leq d_1 d_2$ and for any $\de>0$, the minimizer $\hat{M}_{\max}$ of the optimization program \eq{max-est} satisfies
\be
	\| \hat{M}_{\max} - M^* \|_{\Pi}^2 = \sum_{k=1}^{d_1}\sum_{l=1}^{d_2} \pi_{kl} \{ \hat{M}_{max} - M^* \}_{k,l}^2 \leq  C \beta_{\alpha} \bigg\{  L_{\alpha}R \s{\f{ d}{n}}  + U_{\alpha}\s{\f{\log(4/\de)}{n}} \, \bigg\},  \lbl{mn-up}
\ee
with probability at least $1-\de$. Here $\| \cdot \|_{\Pi}$ denotes the weighted Frobenius norm with respect to $\Pi$, i.e.,
$$
	\| M \|_{\Pi} = \sqrt{ \sum_{k=1}^{d_1}\sum_{l=1}^{d_2} \pi_{kl}M_{k,l}^2 } \ \  \mbox{ for all } M \in \br^{d_1 \times d_2}.
$$
\end{theorem}

\medskip
\begin{remark}{\rm
\noi
\begin{enumerate}
\item[(i)] While using the trace-norm to study this general weighted sampling model, it is common to assume that each row and column is sampled with positive probability (Nagahban and Wainwright, 2012; Klopp, 2012), though in some applications this assumption does not seem realistic. More precisely, assume that there exists a positive constant $\mu \geq 1$ such that
\be
	\pi_{k l } \geq \f{1}{\mu d_1 d_2}, \ \ \mbox{ for all } (k ,l) \in [d_1]\times [d_2].  \lbl{ass1}
\ee
Then, under condition \eq{ass1} and the conditions of Theorem \ref{thm1-1b},
\be
	\f{1}{d_1d_2} \|\hat{M}_{\max} -M^* \|_F^2  \leq  C \mu \beta_{\alpha} \bigg\{  L_{\alpha} R \s{\f{ d}{n}}  + U_{\alpha}\s{\f{\log (d)}{n}} \, \bigg\}  \lbl{mn-up2}
\ee
holds with probability at least $1-4/d$, where $C>0$ denotes an absolute constant.

\item[(ii)] Klopp (2012) studied the problem of standard matrix completion with noise, also in the case of general sampling distribution, using the trace-norm penalized approach. However, the Assumption 1 therein requires that the distribution $\pi_{kl}$ over entries is bounded from above, which is quite restrictive especially in the Netflix problem. It is worth noticing that this upper bound condition on sampling distribution is not required in both results \eq{mn-up} and \eq{mn-up2}.
\end{enumerate}

}
\end{remark}

It is noteworthy that above results are directly comparable to those obtained in the case of approximately low-rank recovery from unquantized measurements, also using max-norm regularized approach \cite{CaiZ12}. Let $Z=(Z_{i,j})$ be a noise matrix consisting of i.i.d. $N(0,\sigma^2)$ entries for some $\sigma>0$, and assume we have observations on a (random) subset $S=\{(i_1, j_1), ..., (i_n, j_n)\}$ of entries of $\tilde{Y} = M^*+ Z$. Cai and Zhou (2013) studied the unquantized problem under a general sampling model using max-norm as a convex relaxation for the rank. In particular, for the max-norm constrained least squares estimator
\be
	\tilde{M}_{\max}  = \argmin_{M \in K_{\max}(\alpha , R) }  \f{1}{n}\sum_{t=1}^n ( \tilde{Y}_{i_t,j_t} - M^*_{i_t, j_t})^2,	 \lbl{max.lse}
\ee
it was shown that for any $\de \in (0, 1)$ and a sample size $2< n\leq d_1 d_2$,
\be
\| \tilde{M}_{\max} - M^* \|_{\Pi}^2 \leq C'  \bigg\{  (\alpha \vee \sigma) R \s{\f{ d}{n}}  +  	 \f{\alpha^2 \log(2/\de)}{n} \, \bigg\} \lbl{TZ12.ubd}
\ee
holds with probability greater than $1-\exp(-d)-\de$, where $C'>0$ is a universal constant.

In 1-bit observations case when $Z_{i,j}\stackrel{i.i.d.}{\sim}N(0,\sigma^2)$, it is equivalent that the function $F$ in model \eq{1b-md} is given by $F(\cdot)=\Phi(\cdot/\sigma)$. According to \eq{prob-ex}, we have
\be
	\| \hat{M}_{\max} - M^* \|_{\Pi}^2  \leq C \exp\Big( \f{\alpha^2}{2\sigma^2} \Big)  \bigg\{ (\alpha+\sigma) R \s{\f{ d}{n}}  +  	 (\alpha +\sigma)^2\s{\f{\log(4/\de)}{n}} \, \bigg\}  \lbl{1-bit.ubd}
\ee
holds with probability at least $1-\de$.

Comparing the upper bounds in \eq{TZ12.ubd} and \eq{1-bit.ubd} and note that $\alpha \vee \sigma \leq \alpha+\sigma \leq 2(\alpha \vee \sigma)$, we see that there is no essential loss of recovery accuracy by discretizing to binary measurements as long as $\f{\alpha}{\sigma}$ is bounded by a constant \cite{DP}. On the other hand, as the signal-to-noise ratio $\f{\alpha}{\sigma} \geq 1$ increases, the error bounds deteriorate significantly. In fact, the case $\alpha \gg \sigma$ essentially amounts to the noiseless setting, in which it is impossible to recover $M^*$ based on any subset of the signs of its entries.

\medskip
\subsection{Information-theoretic lower bounds}

We now establish minimax lower bounds by using information-theoretic techniques. The lower bounds given in Theorem \ref{mn-low} below  show that the rate attained by the max-norm constrained maximum likelihood estimator is optimal up to constant factors.

\medskip
\begin{theorem} \lbl{mn-low}
Assume that $F'(x)$ is decreasing and $\f{F(x)(1-F(x))}{(F'(x))^2}$ is increasing for $x>0$, and let $S$ be any subset of $[d_1] \times [d_2]$ with cardinality $n$. Then, as long as the parameters $(R,\alpha)$ satisfy
\be
	 \max\bigg( 2,  \, \f{4}{(d_1 \vee d_2)^{1/2}} \bigg)   \leq \f{R}{\alpha}  \leq  \f{ ( d_1 \wedge d_2)^{1/2}}{2} , \lbl{r0}
\ee
the minimax risk for estimating $M$ over the parameter space $K_{\max}(\alpha , R)$ satisfies
\be
	\inf_{\hat{M}} \max_{M \in K_{\max}(\alpha ,R)}  \bigg\{ \f{1}{d_1 d_2} \e \| \hat{M} - M \|_F^2 \bigg\} \geq   \f{1}{512}\min\bigg\{ \alpha^2 ,  \, \f{ \s{\beta_{\alpha/2}}}{2} R \s{\f{ d}{n}} \, \bigg\}.   \lbl{minimax-lbd0}
\ee
\end{theorem}

\medskip

\begin{remark}{\rm
In fact, the lower bound \eq{minimax-lbd0} is a special case of the following general result, which will be proved in Sect.~\ref{pf_lbd}.
Let $\ga^*>0$ be the solution of the following equation
\be
	\ga^*  =  \min\bigg\{ \f{1}{2}, \,  \f{R^{1/2}}{\alpha}\bigg(\f{\beta_{(1-\ga^*)\alpha}}{32 }  \cdot
	\f{ d_1 \vee d_2 }{n} \bigg)^{1/4} \, \bigg\} \lbl{ga*}
\ee
and assume that
\be
	 \max\bigg( 2,  \, \f{4}{(d_1 \vee d_2)^{1/2}} \bigg)  \leq \f{R}{\alpha}  \leq   ( d_1 \wedge d_2)^{1/2}\ga^*.  \lbl{r}
\ee
Then the minimax risk for estimating $M$ over the parameter space $K_{\max}(\alpha , R)$ satisfies
\be
	\inf_{\hat{M}} \max_{M \in K_{\max}(\alpha ,R)}  \bigg\{ \f{1}{d_1 d_2} \e \| \hat{M} - M \|_F^2 \bigg\} \geq   \f{1}{512}\min\bigg\{ \alpha^2 ,  \, \f{ \s{\beta_{(1-\ga^*)\alpha}}}{2} R \s{\f{ d}{n}} \, \bigg\}.   \lbl{minimax-lbd}
\ee
To see the existence of  $\gamma^*$ defined above, setting
$$
	h(\ga)=\ga \ \ \mbox{ and } \ \  g(\ga)= \min\bigg\{ \f{1}{2}, \,  \f{R^{1/2}}{\alpha}\bigg(\f{\beta_{(1-\ga)\alpha}}{32 }  \cdot
	\f{ d_1 \vee d_2 }{n} \bigg)^{1/4} \, \bigg\},
$$
then it is easy to see that $h(\ga)$ is strictly increasing and $g(\ga)$ is decreasing for $\ga \in (0, 1)$ with $h(0)=0$ and $g(0)>0$. Therefore, equation \eq{ga*} has a unique solution $\ga^* \in (0, \f{1}{2}]$, i.e. $h(\ga^*)=g(\ga^*)$.
}
\end{remark}

Assume that $\mu$ and $\alpha$ are bounded above by universal constants and let the function $F$ be fixed, so that both $L_{\alpha}$ and $\beta_{\alpha}$ are bounded. Also notice that $ \beta_{(1-\ga^*)\alpha} \geq \beta_{\alpha/2} $ since $\ga^* \leq 1/2$. Then comparing the lower bound \eq{minimax-lbd} with the upper bound \eq{mn-up2} shows that if the sample size $n \geq \f{R^2\beta_{\alpha/2}}{4\alpha^4}(d_1+d_2)$, the optimal rate of convergence is $R\s{\f{d_1+d_2}{n}}$, i.e.
\[
\inf_{\hat{M}} \sup_{M \in K_{\max}(\alpha, R)}   \f{1}{d_1 d_2} \e \| \hat{M} - M \|_F^2   \asymp   R \s{\f{d_1+d_2}{n}},
\]
and the max-norm constrained maximum likelihood estimate \eqref{max-est} is rate-optimal. If the target matrix $M^*$ is known to have rank at most $r$, we can take $R=\alpha \s{r}$, such that the requirement here on the sample size $n  \geq  \f{\beta_{\alpha/2}}{4\alpha^2} r (d_1+d_2)$ is weak and the optimal rate of convergence becomes $\alpha\s{\f{r(d_1+d_2)}{n}}$.

\subsection{Comparison to prior work}

In this paper, we study a matrix completion model proposed in \cite{DP}, in which it is assumed that a binary matrix is observed at random from a distribution parameterized by an unknown matrix which is (approximately) low-rank. It is noteworthy that some earlier papers on collaborative filtering or matrix completion, including Srebro, {\it et al.} (2004) and references therein, also dealt with binary observations that are assumed to be noisy versions of the underlying matrix, in Logistic or Bernoulli conditional model. The goal there is to predict directly the quantized values, or equivalently, to reconstruct the sign matrix, instead of the underlying real values, therefore the non-identifiability issue could be avoided.

We next turn to a detailed comparison of our results for 1-bit matrix completion to those obtained in \cite{DP}, also for approximately low-rank matrices. Using the trace-norm as a proxy to rank, Davenport, \textit{et al.} (2012) have studied 1-bit matrix completion under the {\it uniform sampling distribution} over the parameter space
$$
	K_*(\alpha, r)  = \Big\{ M \in \br^{d_1 \times d_2}: \| M \|_{\infty} \leq \alpha, \;  \f{\| M\|_* }{\s{d_1 d_2}} \leq \alpha \s{r} \Big\},
$$
for some $\alpha>0$ and $r\leq \min\{d_1, d_2\}$ is a positive integer. To recover the unknown $M^* \in K_*(\alpha, r)$, given a collection of observations $Y_S$ where $S$ follows a Bernoulli model, i.e. every entry $(k,l) \in [d_1] \times [d_2]$ is observed 
independently with equal probability $\f{n}{d_1 d_2}$, they propose the following trace-norm constrained MLE
\be
	\hat{M}_{\rm tr} =  \argmin_{M \in K_*(\alpha , r)} \, \ell_S(M;Y)  \lbl{tn-est}
\ee
and prove that for a sample size $n \geq d \log(d)$, $d=d_1+d_2$, with high probability,
\be
	\f{1}{d_1 d_2} \| \hat{M}_{\rm tr} -M^* \|_F^2 \lesssim  \beta_{\alpha} L_{\alpha} \alpha \s{\f{r d}{n}}. \lbl{DP-ubd}
\ee
Comparing to \eq{mn-up2} with $R=\alpha \s{r}$, it is easy to see that under the uniform sampling model, the error bounds in (rescaled) Frobenius norm for the two estimates $\hat{M}_{\max}$ and $\hat{M}_{\rm tr}$ are of the same order. Moreover, Theorem 3 in \cite{DP} and Theorem \ref{mn-low}, respectively, provide lower bounds showing that both $\hat{M}_{\rm tr}$ and $\hat{M}_{\max}$ achieve the minimax rate of convergence for recovering approximately low-rank matrices over the parameter spaces $K_*(\alpha, r)$ and $K_{\max}(\alpha, R)$ respectively.

As mentioned in the introduction, the uniform sampling distribution assumption is  restrictive and  not valid in many applications including the well-known Netflix problem. When the sampling distribution is non-uniform, it was shown in Salakhutdinov and Srebro (2010) that the standard trace-norm regularized method might fail, specifically in the setting where the row and column marginal distributions are such that certain rows or columns are sampled with very high probabilities. Moreover, it was proposed to use a weighted variant of the trace-norm, which incorporates the knowledge of the true sampling distribution in its construction, and showed experimentally that this variant indeed leads to superior performance. Using this weighted trace-norm, Negahban and Wainright (2012) provided theoretical guarantees on approximate low-rank matrix completion in general sampling case while assuming that each row and column is sampled with positive probability (See condition \eq{ass1}). In addition, requiring that the probabilities to observe an element from any row or column are of order $O((d_1 \wedge d_2)^{-1})$, Klopp (2012) analyzed the performance of the trace-norm penalized estimators, and provided near-optimal (up to a logarithmic factor) bounds  which are similar to the bounds in this paper.

Next we provide an analysis of the performance of the weighted trace-norm in 1-bit matrix completion. Given the knowledge of the true sampling distribution, we establish an upper bound on the error in recovering $M^*$, which comparing to \eq{DP-ubd}, includes an additional $\log^{1/2}(d)$ factor. We do not rule out the possibility that this logarithmic factor might be an artifact of the technical tools used in proof described below. The proof in \cite{DP} for the trace-norm regularization in uniform sampling case may also be extended to the weighted trace-norm method under the general sampling model, by using the matrix Bernstein inequality instead of Seginer's theorem. The extra logarithmic factor, however, is still inevitable based on this argument. We will not pursue the details in this paper.

Given a sampling distribution $\Pi=\{\pi_{kl}\}$ on $[d_1] \times [d_2]$, define its row- and column-marginals as
$$
	\pi_{k \cdot} = \sum_{l=1}^{d_2} \pi_{kl} \ \ \mbox{ and } \ \
	\pi_{\cdot l} =\sum_{k=1}^{d_1} \pi_{kl},
$$
respectively. Under the condition \eq{ass1}, we have
\be
	\pi_{k \cdot } \geq \f{1}{\mu d_1}, \ \ \pi_{\cdot l} \geq \f{1}{\mu d_2}, \ \ \mbox{ for all } (k,l) \in [d_1] \times [d_2].	\lbl{ass2}
\ee
As in \cite{SS10}, consider the following weighted trace-norm with respect to the distribution $\Pi$:
\be
	\| M \|_{w, *}  := \| M_w \|_* = \big\| \mbox{diag}(\s{\pi_{1 \cdot }}, ..., \s{\pi_{d_1 \cdot }})
	 \cdot M  \cdot \mbox{diag}(\s{\pi_{ \cdot 1}}, ..., \s{\pi_{ \cdot d_2}}) \big\|_* ,  \lbl{wtn}
\ee
where $(M_w)_{k,l} := \s{\pi_{k \cdot } \pi_{\cdot l}} M_{k,l}$. Notice that if $M$ has rank at most $r$ and $\|M \|_{\infty} \leq \alpha$, then
\beq
	\| M \|_{w,*} \leq \s{r} \| M \|_F = \s{r}\bigg( \sum_{k=1}^{d_1} \sum_{l=1}^{d_2} \pi_{k \cdot }\pi_{\cdot l} M_{k,l}^2 \bigg)^{1/2}  \leq \alpha \s{r}.
\eeq
Analogous to the previous studied class $K_*(\alpha, r)$ containing the low trace-norm matrices, define
\beq
	K_{\Pi, *}  \equiv K_{\Pi,*}( r, \alpha )  =\Big\{  M \in \br^{d_1 \times d_2}: \| M \|_{w,*} \leq \alpha \s{r},  \| M \|_{\infty} \leq \alpha \Big\}
\eeq
and consider estimating the unknown $M^* \in K_{\Pi, *}$ by solving the following optimization problem:
\be
	 \hat{M}_{w,tr}  = \argmin_{M \in K_{\Pi,*}} \ell_S(M; Y).  \lbl{weighted}
\ee
The following theorem states that the weighted trace-norm regularized approach can be nearly as good as the max-norm regularized estimator (up to logarithmic and constant factors),  under a general weighted sampling distribution. The theoretical performance of the weighted trace-norm is first studied by Foygel, {\it et al.} (2011) in the standard matrix completion problems under arbitrary sampling distributions.

\begin{theorem} \lbl{wei}
Suppose that {\rm Condition U} holds but with $M^* \in  K_{\Pi,*}$, assume that the training set $S$ follows a general weighted sampling model according to the distribution $\Pi$ satisfying \eq{ass1}. Then there exists an absolute constant $C>0$ such that, for a sample size $n \geq \mu \min\{ d_1, d_2\}\log(d)$ and any $\de>0$, the minimizer $\hat{M}_{w, tr}$ of the optimization program \eq{weighted} satisfies
\be
	\| \hat{M}_{w,tr} - M^* \|_{\Pi}^2  \leq  C  \beta_{\alpha} \bigg\{   L_{\alpha} \alpha \s{\f{\mu r d\log(d)}{n}}
	+  U_{\alpha}\s{\f{\log(4/\de)}{n}}  \bigg\}, \lbl{wei-up}
\ee
with probability at least $1-\de$.
\end{theorem}

Since the construction of weighted trace-norm $\| \cdot \|_{w, *}$ highly depends on the underlying sampling distribution which is typically unknown in practice, the constraint $M^* \in  K_{\Pi,*}$ seems to be artificial. The max-norm constrained approach, on the contrary, does not require the knowledge of the exact sampling distribution and the error bound in weighted Frobenius norm, as shown in \eq{mn-up}, holds even without prior assumption on $\Pi$, e.g., \eq{ass1}.

To clarify the major difference between the principles behind \eq{DP-ubd} and \eq{wei-up}, we remark that one of the key technical tools used in \cite{DP} is a bound of Seginer (2000) on the spectral norm of a random matrix with i.i.d. zero mean entries (corresponding to the uniform sampling distribution), i.e. for any $h \leq 2\log(\max\{d_1, d_2\})$,
\beq
	\e [\| A \|^h] \leq K^h \Big( \e \Big[ \max_{k=1, ..., d_1} \| a_{k \cdot } \|_2^h \Big] + \e \Big[\max_{j=1, ..., d_2} \| a_{\cdot l} \|_2^h \Big]  \Big),
\eeq
where $a_{k\cdot}$ (resp. $a_{\cdot l}$) denote the rows (resp. columns) of $A$ and $K$ is a universal constant. Under the non-uniform sampling model, we will deal with a matrix with independent entries that are not necessarily identically distributed, to which case an alternative result of Latala (2005) can be applied, i.e.
\beq
	\e [\| A \|] \leq K' \Big( \max_{k=1, ..., d_1} \e  \| a_{k \cdot } \|_2 +  \max_{j=1, ..., d_2} \e \| a_{\cdot l} \|_2 + \Big( \sum_{k,l}\e a_{kl}^4 \Big)^{1/4} \Big]  \Big),
\eeq
or instead, resorting to the matrix Bernstein inequality. Using either inequality would thus bring an additional logarithmic factor, appeared in \eq{wei-up}.

It is also worth noticing that though the sampling distribution is not known exactly in practice, its empirical analogues are expected to be stable enough as an alternative. According to Forgel, {\it et al.} (2011), given a random sample $S=\{(i_t,j_t)\}_{t=1}^n$, consider the empirical marginals
$$
	\hat{\pi}^r(i) = \f{\# \{t:i_t=i \}}{n}, \ \ \hat{\pi}^c(j)=\f{\# \{t: j_t = j\}}{n}
	\ \ \mbox{ and } \ \  \hat{\pi}_{ij} = \hat{\pi}^r(i)\hat{\pi}^c(j),
$$
as well as the smoothed empirical marginals
$$
	\check{\pi}^r(i) = \f{1}{2}(\hat{\pi}^r(i) + 1/d_1), \ \ \hat{\pi}^c(j)=\f{1}{2}(\hat{\pi}^c(j)+1/d_2)
	\ \ \mbox{ and } \ \  \check{\pi}_{ij} = \check{\pi}^r(i)\check{\pi}^c(j).
$$
The smoothed empirically-weighted trace-norm $\| \cdot \|_{\check{w},*}$ can be defined in the same spirit as in the definition \eq{wtn} of weighted trace-norm, only with $\{\pi_{ij}\}$ replaced by $\{\check{\pi}_{ij}\}$. Then the unknown matrix can be estimated via regularization on the $\check{\pi}$-weighted trace-norm, that is,
$$
\check{M}_{\check{w},tr} = \argmin\big\{ \ell_S(M; Y):  \| M \|_{\infty} \leq \alpha, \;  \| M \|_{\check{w},*} \leq \alpha \s{r} \big\}.
$$
Adopting \cite[Theorem 4]{FSSS} to the current 1-bit problem will lead to a learning guarantee similar to \eq{wei-up}.

\section{Computational Algorithm}
\lbl{implementation.sec}
\setcounter{equation}{0}

Problems of the form \eq{max-est} can now be solved using a variety of algorithms, including interior point method \cite{Sre04}, Frank-Wolfe-type algorithm \cite{MJ13} and projected gradient method \cite{Lee2010}. The first two are convex methods with guaranteed convergence rates to the global optimum, though can be slow in practice and might not scale to matrices with hundreds of rows or columns. We describe in this section a simple first order method due to Lee, {\it et al.} (2010), which is a special case of a projected gradient algorithm for solving large-scale convex programs involving the max-norm. This method is non-convex, but as long as the size of the problem is large enough, it is guaranteed that each local minimum is also a global optimum, due to Burer and Monteiro (2003).

We start from rewriting the original problem as an optimization over factorizations of a matrix $M \in \br^{d_1 \times d_2}$ into two terms $M= U V^T$, where $U \in \br^{d_1 \times k}$ and $V \in \br^{d_2 \times k}$ for some $1\leq k \leq d=d_1+d_2$. More specifically, for any $1\leq k\leq d$ fixed, define
\beq
	\mathcal{M}_k(R) := \Big\{  U V^T :  U \in \br^{d_1 \times k}, \,  V \in \br^{d_2 \times k}, \max\{\| U \|_{2,\infty}^2, \| V \|_{2,\infty}^2 \} \leq  R \Big\}.
\eeq
Then the global optimum of \eq{max-est} is equal to that of
\beqn
\mbox{minimize} && \ell(M;Y)   \ \   \nn \\
\mbox{ subject to} &&  M \in \mathcal{M}_k(R),\ \  \| M \|_{\infty}\leq \alpha.   \label{est2'}
\eeqn
Here we write $\ell(M;Y) = \f{1}{|S|} \ell_S(M;Y)$ for brevity. This problem is non-convex, come with no guaranteed convergence rates to the global optimum. A surprising fact is that when $k\geq 1$ is large enough, this problem has no local minimum \cite{BM03}. Notice that $\ell(\cdot; Y)$ is differentiable with respect to the first argument, then \eq{est2'} can be solved iteratively via the following updates:
	\[
	\left[\begin{array}{c}
	U(\tau)  \\
	V(\tau)
	\end{array}\right]    =  \left[\begin{array}{c}
	U^{t} - \f{\tau}{\sqrt{t}  } \cdot  \nabla f(U^t (V^{t })^T; Y )V^t \\
	V^{t} - \f{\tau}{\sqrt{t}} \cdot  \nabla f(U^t (V^{t })^T; Y )^T U^t
	\end{array}\right],
\]
where $\tau>0$ is a stepsize parameter and $t=0, 1, 2, ...$. Next, we project $(U(\tau),V(\tau))$ onto $\mathcal{M}_k(R)$ according to
\[
	  \left[\begin{array}{c}
	\tilde{U}^{t+1}  \\
	\tilde{V}^{t+1}
	\end{array}\right]   = \mathcal{P}_R \bigg( \left[\begin{array}{c}
	U(\tau)  \\
	V(\tau)
	\end{array}\right]  \bigg).
\]
This orthogonal projection can be computed by re-scaling the rows of the current iterate whose $\ell_2$-norms exceed $R$ so that their norms become exactly $R$, while rows with norms already less than $R$ remain unchanged. If $\| \tilde{U}^{t+1} (\tilde{V}^{t+1})^T \|_{\infty}> \alpha$, we replace
\[
	\left[\begin{array}{c}
	\tilde{U}^{t+1}  \\
	\tilde{V}^{t+1}
	\end{array}\right]   \ \ \mbox{ with } \ \  \f{\s{\alpha}}{\| \tilde{U}^{t+1} (\tilde{V}^{t+1})^T \|_{\infty}^{1/2}} \left[\begin{array}{c}
	\tilde{U}^{t+1}  \\
	\tilde{V}^{t+1}
	\end{array}\right],
\]
otherwise we keep it still. The resulting update is then denoted by $(U^{t+1},V^{t+1})$.

It is important to note that the choice of $k$ must be large enough, at least as big as the rank of $M^*$. Suppose that, before solving \eq{max-est}, we know that the target matrix $M^*$ has rank at most $r^*$. Then it is best to solve \eq{est2'} for $k=r^*+1$ in the sense that, if we choose $k \leq r^*$, then \eq{est2'} is not equivalent to \eq{max-est}, and if we take $k > r^* +1$, then we would be solving a larger program than necessary. In practice, we do not know the exact value of $r^*$ in advance. Nevertheless, motivated by Burer and Monteiro (2003), we suggest the following scheme to solve the problem which avoids solving \eq{est2'} for $r \gg r^*$:

\begin{enumerate}
\item[(1)] Choose an initial small $k$ and compute a local minimum $(U,V)$ of \eq{est2'}, using above projected gradient method.

\item[(2)] Use an optimization technique to determine whether the injections $\hat{U}$ of $U$ into $\br^{d_1 \times (k+1)}$ and $\hat{V}$ of $V$ into $\br^{d_2 \times (k+1)}$ comprise a local minimum of \eq{est2'} with the size increased to $k+1$.

\item[(3)] If $(\hat{U},\hat{V})$ is a local minimum, then we can take $M=UV^T$ as the final solution; otherwise compute a better local minimum $(\tilde{U},\tilde{V})$ of \eq{est2'} with size $k+1$ and repeat step (2) with $(U,V)=(\tilde{U},\tilde{V})$ and $k=k+1$.
\end{enumerate}

It was also suggested in \cite{Lee2010} that when dealing with extremely large datasets with $S$ consisting of hundreds of millions of index pairs, one may consider using a stochastic gradient method based on the following decomposition for $\ell$, that is,
\beq
	\ell(UV^T; Y) &= & \f{1}{|S|}\sum_{(i,j) \in S }
g(u_{i}^T v_{j}; Y_{i , j}) \ \ \mbox{ with } \\
 g(t;y) & = & \textbf{1}_{\{y=1\}}\log\bigg(\f{1}{ F(t)}\bigg)  + \textbf{1}_{\{y=-1\}}\log\bigg(\f{1}{1-F(t)}\bigg),
\eeq
where $S  \subset [d_1] \times [d_2]$ is a training set of row-column indices, $u_i$ and $v_j$ denote the $i$-th row of U and $j$-th row of V, respectively. The stochastic gradient method says that at $t$-th iteration, we only need to pick one training pair $(i_t,j_t)$ at random from $S$, then update $g(u_{i_t}^T v_{j_t}; Y_{i_t, j_t})$ via the previous procedure. More precisely, if $\| u_{i_t} \|_2^2 > R$, we project it back so that $\| u_{i_t} \|_2^2 = R$, otherwise we do not make any change (do the same for $v _{j_t}$). Next, if $| u_{i_t}^T v_{j_t}|>\alpha$, replace $u_{i_t}$ and $v_{i_t}$ with $\s{\alpha}u_{i_t}/|u_{i_t}^T v_{j_t}|^{1/2}$ and $\s{\alpha}v_{i_t}/|u_{i_t}^T v_{j_t}|^{1/2}$ respectively, otherwise we keep everything still. At the $t$-th iteration, we do not need to consider any other rows of $U$ and $V$. This simple algorithm could be computationally as efficient as optimization with the trace-norm.

\section{Numerical results}
\lbl{sim.sec}
\setcounter{equation}{0}

In this section, we report the simulation results for low-rank matrix recovery based on 1-bit observations. In all cases presented below, we solved the convex program \eq{est2'} by using our implementation in MATLAB of the projected gradient algorithm proposed in Sect.~\ref{implementation.sec} for a wide range of values of  the step-size parameter $\tau$.

We first consider a rank-$2$, $d\times d$ target matrix $M^*$ with eigenvalues $\{d/\sqrt{2}, d/\sqrt{2}, 0, ..., 0\}$, so that $\| M^* \|_F/d=1$. We choose to work with the Gaussian conditional model under uniform sampling. Let $Y_S$ be the noisy binary observations with $S=\{(i_1, j_1), ..., (i_t, j_t)\}$, that is,
for $(i,j)\in S$,
\beq
	Y_{i, j}  = \left\{\begin{array}{ll}
	+1,    & \mbox{ with probability } \Phi(M^*_{i,j}/\sigma),  \\
    -1,    &  \mbox{ with probability } 1- \Phi(M^*_{i,j}/\sigma),
					\end{array}  \right.
\eeq
and the objective function is given by
\beq
	\ell_S(M;Y)=  \frac{1}{|S|} \bigg\{ \sum_{(i,j)\in \Omega^+}
	\log\Big[ \frac{1}{\Phi(M_{i,j}/\sigma)} \Big] + \sum_{(i,j)\in \Omega^-}
	\log\Big[\frac{1}{1-\Phi(M_{i,j}/\sigma)} \Big] \bigg\},
\eeq
where $\Omega^+ = \{(i,j)\in S: Y_{i,j}=1\}$ and $\Omega^- = \{(i,j)\in S: Y_{i,j}= -1\}$. In Figure 1, averaging the results over 20 repetitions, we plot the squared Frobenius norm of the error (normalized by the dimension) $\| \hat{M}-M^*\|_F^2/d^2$ versus a range of sample sizes $s=|S|$, with the noise level $\sigma$ taken to be $\alpha/2$, for three different matrix sizes, $d\in \{80,120,160\}$. Naturally, in each case, the Frobenius error decays as $s$ increases, although larger matrices require larger sample sizes, as reflected by the upward shift of the curves as $d$ is increased.

\begin{figure}
\centering
\includegraphics[scale=0.8]{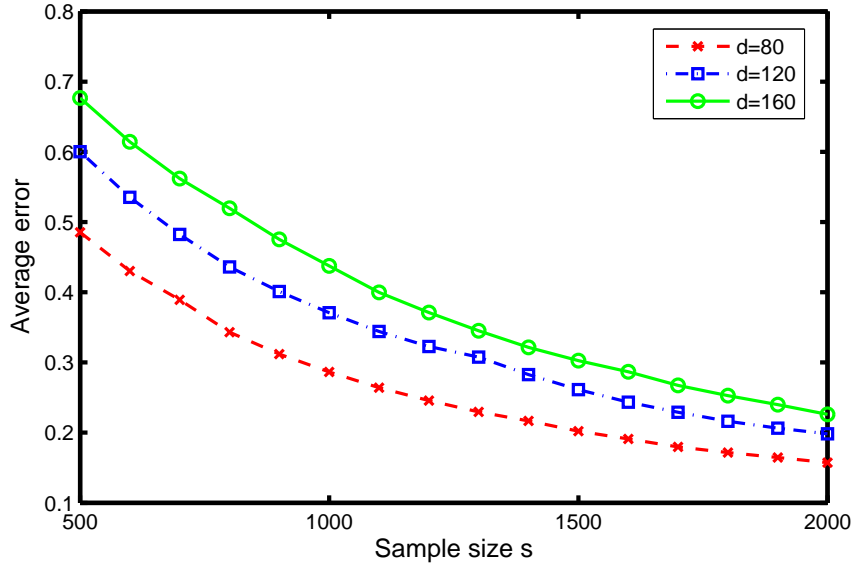}
\caption{Plot of the average Frobenius error $\| \hat{M}-M^*\|_F^2/d^2$ versusu the sample size s for three different matrix sizes $d\in\{80,120,160\}$, all with rank $r=2$.}
\end{figure}

Next, we compare the performance of the max-norm based regularization with that of the trace-norm using the same criterion as in \cite{DP}. More specifically, the target matrix $M^*$ is constructed at random by generating $M = L R^T$, where $L$ and $R$ are $d\times r$ matrices with i.i.d. entries drawn from Uniform~$[-1/2,1/2]$, so that rank$(M^*)=r$. It is then scaled such that $\| M^* \|_{\infty}=1$, while in the last case, $M^*$ is formed such that $\|M^*\|_F/d=1$. As before, we focus on the Gaussian conditional model but with noise level $\sigma$ varies from $10^{-3}$ to $10$, and set $d=500$, $r=1$ and $s=0.15 d^2$, which is exactly the same case studied in \cite{DP}. We plot in Figure 2 the squared Frobenius norm of the error (normalized by the norm of the underlying matrix $M^*$) over a range of different values of noise level $\sigma$ on a logarithmic scale. As evident in Figure 2, the max-norm based regularization performs slightly but consistently better than the trace-norm, except on the one point where $\sigma=\log_{10}(0.25)$. Also, we see that for both methods, the performance is poor when the noise is either too little or too much.

\begin{figure}
\centering
\includegraphics[scale=0.8]{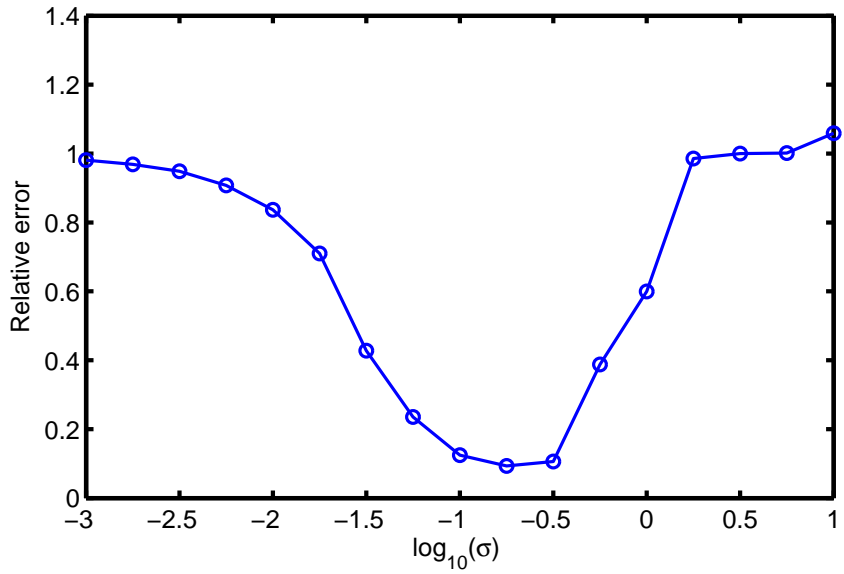}
\caption{Plot of the relative Frobenius error $\|\hat{M} -M^*\|_F^2/\|M^*\|_F^2$ versus the noise level $\sigma$ on a logarithmic scale, with rank $r=1$.}
\end{figure}

In the third experiment, we consider matrices with dimension $d=200$ and choose a moderate level of noise, that is, $\sigma=\log_{10}(-0.75)$, according to previous experiences. Figure 3 plots the relative Frobenius norm of the error versus the sample size $s$ for three different matrix ranks, $r\in \{3,5,10\}$. Indeed, larger rank means larger intrinsic dimension of the problem, and thus increases the difficulty of any reconstruction procedure.

\begin{figure}
\centering
\includegraphics[scale=0.8]{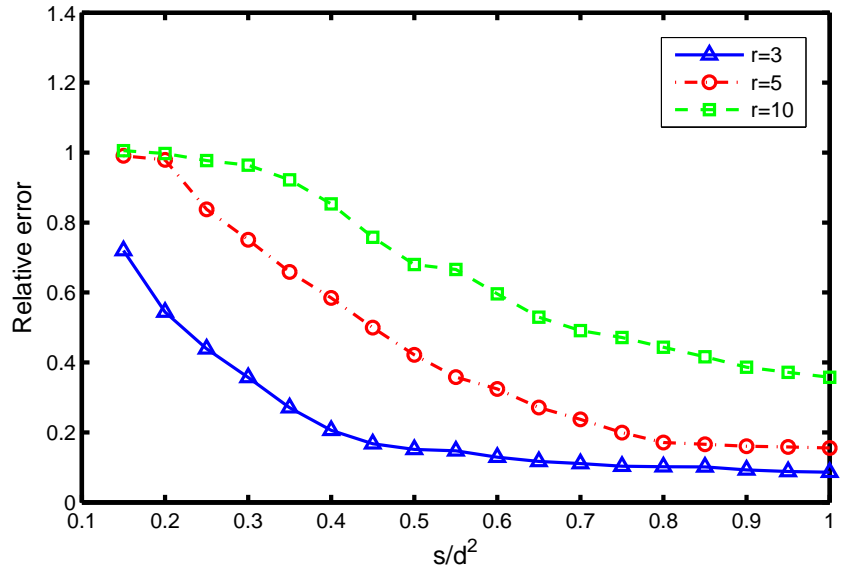}
\caption{Plot of the relative Frobenius error versus the rescaled sample size $s/d^2$ for three different ranks $r\in \{3,5, 10\}$, all with matrix size $d=200$.}
\end{figure}

\section{Discussion}
\label{dscs.sec}
\setcounter{equation}{0}

This paper studies the problem of recovering a low-rank matrix based on highly quantized (to a single bit) noisy observation of a subset of entries. The problem was first formulated and studied by Davenport, {\it et al.} (2012), where the authors consider approximately low-rank matrices in terms that the singular values belong to a scaled Schatten-1 ball. When the infinity norm of the unknown matrix $M^*$ is bounded by a constant and its entries are observed uniformly in random,  they show that $M^*$ can be recovered from binary measurements accurately and efficiently.

Our theory, on the other hand, focuses on approximately low-rank matrices in the sense that unknown matrix belongs to certain max-norm ball. The unit max-norm ball is nearly the convex hull of rank-1 matrices whose entries are bounded in magnitude by 1, thus is a natural convex relaxation of low-rank matrices, particularly with bounded infinity norm. Allowing for non-uniform sampling, we show that the max-norm constrained maximum likelihood estimation is rate-optimal up to a constant factor, and that the corresponding convex program may be solved efficiently in polynomial time. An interesting question naturally arises that whether it is possible to push the theory further to cover exact low-rank matrix completion from noisy binary measurements.

In our previous work \cite{CaiZ12}, we suggest to use max-norm constrained least square estimation to study standard matrix completion (based on noisy observations) under a general sampling model. Similar errors bounds are obtained, which are tight to within a constant. Comparing both results in the case of Gaussian noise demonstrates that as long as the signal-to-noise ratio remains constant, almost nothing is lost by quantizing to a single bit.

\section{Proofs}
\label{proof.sec}
\setcounter{equation}{0}

\subsection{Proof of Theorem \ref{thm1-1b}}

The proof of Theorem \ref{thm1-1b} is based on general excess risk bounds developed in Bartlett and Mendelson (2002) for empirical risk minimization when the loss function is Lipschitz. We regard matrix recovery as a prediction problem, that is, consider a matrix $M \in \br^{d_1 \times d_2}$ as a function: $[d_1] \times [d_2] \rightarrow \br$, i.e. $M(k,l)=M_{k,l}$. Moreover, define a function $g(x;y)$ $\br\times \{ \pm 1\} \mapsto \br$, which can be seen as a loss function:
\beq
	g(x;y)=  \textbf{1}_{\{y=1\}}\log\bigg(\f{1}{ F(x)}\bigg)  + \textbf{1}_{\{y=-1\}}\log\bigg(\f{1}{1-F(x)}\bigg).
\eeq
For a subset $S =\{(i_1, j_1), ... , (i_n, j_n) \} \subseteq ( [d_1] \times [d_2] )^n$ of the observed entries of $Y$, let $\mathcal{D}_S(M;Y)=\f{1}{n}\sum_{t=1}^n g(M_{i_t, j_t}; Y_{i_t, j_t}) = \f{1}{n} \ell_S(M;Y)$ be the average empirical likelihood function, where the training set $S$ is drawn i.i.d. according to $\Pi$ (with replacement) on $[d_1] \times [d_2]$. Then we have
\beq
	\mathcal{D}_{\Pi}(M;Y) : = \e_{S \sim \Pi}[g(M_{i_t, j_t};Y_{i_t, j_t})] = \sum_{(k,l)\in [d_1] \times [d_2]} \pi_{kl}  \cdot g(M_{k, l};Y_{k,l}).
\eeq
Under condition (U3), we can consider $g$ as a function: $[-\alpha, \alpha] \times \{\pm 1\} \rightarrow \br$, such that for any $y \in  \{\pm 1\}$ fixed, $g(\cdot ; y)$ is essentially an $L_{\alpha}$-Lipschitz loss function. Also notice that in the current case, $Y_{i,j}$ take $\pm 1$ values and appear only in indicator functions, $\textbf{1}\{Y_{i,j}=1\}$ and $\textbf{1}\{Y_{i,j}=-1\}$. Therefore, a combination of Theorem 8,  (4) of Theorem 12 from \cite{BM2001} as well as the upper bound \eq{erc-bd} on the Rademacher complexity of the unit max-norm ball yields that, for any $\de>0$, the following inequality holds with probability at least $1-\de$ over choosing a training set $S$ of $2< n\leq d_1 d_2$ index pairs according to $\Pi$:
\beqn
	\lefteqn{ \sup_{M \in K_{\max}(\alpha , R)} \big(  \e_Y \mathcal{D}_{\Pi}(M;Y) - \e_Y \mathcal{D}_S(M;Y) \big)
	} \nn \\
	&&  \leq 17L_{\alpha}  R\s{\f{ d  }{n}}  + U_{\alpha}\s{\f{8 \log(2/\de)}{n}} := R_n(\alpha, r; \de).   \lbl{riskbd1}
\eeqn
Since $\hat{M}_{\max}$ is optimal and $M^*$ is feasible to the optimization problem \eq{max-est}, we have
$$
	\mathcal{D}_S(\hat{M}_{\max};Y) \leq  \mathcal{D}_S(M^*;Y)  = \f{1}{n} \sum_{t=1}^n g(M^*_{i_t,j_t};Y_{i_t,j_t}).
$$
Since $M^*$ has a fixed value which does not depend on $S$, the empirical likelihood term $\mathcal{D}_S(M^*;Y)$ is an unbiased estimator of $\mathcal{D}_{\Pi}(M^*;Y)$, i.e.
$$
	\e_{S \sim \Pi}[\mathcal{D}_S(M^*;Y)] = \mathcal{D}_{\Pi}(M^*;Y).
$$
However, we still need to find an upper bound on the deviation $\mathcal{D}_S(M^*;Y) - \mathcal{D}_{\Pi}(M^*;Y) $ that holds with high probability. Now, let $A_1, ..., A_n$ be independent random variables taking values in $[d_1] \times [d_2]$ according to $\Pi$, that is, $\p[A_t=(k,l)]=\pi_{kl}$, $t=1, ..., n$, such that $\mathcal{D}_S(M^*;Y)=\f{1}{n}\sum_{t=1}^n g(M_{A_t}; Y_{A_t})$ and
\beq
	\mathcal{D}_S(M^*;Y)- \mathcal{D}_{\Pi}(M^*;Y) = \f{1}{n} \sum_{t=1}^n \big( g(M^*_{A_t};Y_{A_t})- \e[g(M^*_{A_t};Y_{A_t})] \big).
\eeq
Then we will apply the Hoeffding's inequality to the random variables $Z_{A_t}:= g(M^*_{A_t};Y_{A_t})- \e[g(M^*_{A_t};Y_{A_t})]$, conditionally on $Y$. To this end, observe that $0\leq g(M^*_{A_t};Y_{A_t}) \leq U_{\alpha}$ almost surely for all $1\leq t\leq n$, thus for any positive $t$, we have
\be
	 \P_{S \sim \Pi} \big\{  \mathcal{D}_S(M^*;Y) - \mathcal{D}_{\Pi}(M^*;Y)  > t \big\}  \leq   \exp\Big(-\f{2nt^2}{U_{\alpha}^2} \Big),  \lbl{concentration}
\ee
which in turn implies that that with probability at least $1-\de$ over choosing a subset $S$ according to $\Pi$,
\be
	\mathcal{D}_S(M^*;Y)- \mathcal{D}_{\Pi}(M^*;Y) \leq  U_{\alpha} \s{\f{\log (1/\de)}{2n}}.   \lbl{up-2}
\ee
Putting pieces together, we get
\beqn
	\lefteqn{ \e_Y \big[ \mathcal{D}_{\Pi}(\hat{M}_{\max};Y) -\mathcal{D}_{\Pi}(M^*;Y) \big] } \nn \\
	 &= & \e_Y\big[ \mathcal{D}_{\Pi}(\hat{M}_{\max};Y)
	-\mathcal{D}_S(M^*;Y) \big] + \e_Y \big[ \mathcal{D}_S(M^*;Y) - \mathcal{D}_{\Pi}(M^*;Y)  \big]   \nn \\
	&\leq & \e_Y\big[ \mathcal{D}_{\Pi}(\hat{M}_{\max};Y)
	-\mathcal{D}_S(\hat{M}_{\max};Y) \big]
	+ \e_Y \big[ \mathcal{D}_S(M^*;Y) - \mathcal{D}_{\Pi}(M^*;Y)  \big]  \nn \\
	&\leq &    \sup_{M \in K_{\max}(\alpha ,R )} \big\{  \e_Y [ \mathcal{D}_{\Pi}(M;Y) ] -
	\e_Y  [ \mathcal{D}_S(M;Y) ] \big\}   \lbl{up-1}  \\
	&&  +  \e_Y \big[ \mathcal{D}_S(M^*;Y) - \mathcal{D}_{\Pi}(M^*;Y)  \big]. \nn
\eeqn
Moreover, observe that the left-hand side of \eq{up-1} is equal to
\beq
		\lefteqn{   \e_Y \big[ \mathcal{D}_{\Pi}(\hat{M}_{\max};Y) - \mathcal{D}_{\Pi}(M^*;Y) \big] }  \\
	&=& \sum_{(k,l)\in [d_1]\times [d_2]}  \pi_{kl} \bigg[ F(M^*_{k,l})\log\bigg( \f{F(M^*_{k,l})}{F( (\hat{M}_{\max})_{k,l})}\bigg) + 	 (\bar{F}(M^*_{k,l}))  \log\bigg( \f{\bar{F}(M^*_{k,l})}{\bar{F}(  (\hat{M}_{\max})_{k,l})}\bigg)  \bigg],
\eeq
which is the weighted Kullback-Leibler divergence between matrices $F(M)$ and $F(\hat{M}_{\max})$, denoted by $\mathbb{K}_{\Pi}(F(M) \|  F(\hat{M}_{\max}) )$, where
$$
	\bar{F}(\cdot) :=1-F(\cdot )\ \  \mbox{ and } \ \   F(M) := (F(M_{k,l}))_{d_1 \times d_2}.
$$
This, combined with \eq{riskbd1}, \eq{up-2} and \eq{up-1} implies that for any $\de>0$, the following inequality holds with probability at least $1-\de$ over $S$:
\beq
	\mathbb{K}_{\Pi}(F(M^*) \|  F(\hat{M}_{\max}))  \leq   R_n(\alpha, r; \de/2) + U_{\alpha}\s{\f{\log(2/\de)}{2n}}.
\eeq
This, together with \eq{dis} and Lemma \ref{lm2} below gives \eq{mn-up}.

\medskip
\begin{lemma}[Lemma 2, \cite{DP}]  \lbl{lm2}
Let $F$ be an arbitrary differentiable function, and $s, t$ are two real numbers satisfying $|s|, |t| \leq \alpha$. Then
\beq
	d_H^2(F(s); F(t)) \geq \inf_{|x|\leq \alpha} \f{ (F'(x))^2 }{8F(x)(1-F(x))} \cdot  (s-t)^2
\eeq
\end{lemma}

The proof of Theorem \ref{thm1-1b} is now completed. \bbox \\

\subsection{Proof of Theorem \ref{mn-low}}
\lbl{pf_lbd}

The proof for the lower bound follows an information-theoretic method based on Fano's inequality \cite{fano}, as used in the proof of Theorem 3 in \cite{DP}. To begin with, we have the following lemma which guarantees the existence of a suitably large packing set for $K_{\max}(\alpha, R)$ in the Frobenius norm. The proof follows from Lemma 3 of \cite{DP} with a simple modification, see, e.g., the proof of Lemma 3.1 in \cite{CaiZ12}.

\medskip
\begin{lemma}  \lbl{cover}
Let $r=(R/\alpha)^2$ and $\ga \leq 1$ be such that $\f{r}{ \ga^2} \leq \min(d_1, d_2)$ is an integer. There exists a subset $\mathcal{S}(\alpha, \ga) \subset K_{\max}(\alpha, R)$ with cardinality
$$
	 | \mathcal{S}(\alpha, \ga) |  = \bigg[ \exp\bigg( \f{ r \max(d_1, d_2) }{16 \ga^2} \bigg) \bigg] +1
$$
and with the following properties:
\begin{enumerate}
\item[\rm{(i)}] For any $N \in \mathcal{S}(\alpha, \ga)$, \rm{rank}$(N)\leq \f{r}{\ga^2}$ and $N_{k,l} \in \{ \pm \ga
\alpha/2 \}$, such that
$$
		\| N \|_{\infty} =  \f{\ga \alpha}{2} , \ \  \f{1}{d_1 d_2}\| N \|_F^2 =  \f{\ga^2 \alpha^2}{4}.
$$
\item[\rm{(ii)}] For any two distinct $N^k, N^l \in \mathcal{S}(\alpha, \ga)$,
\beq
	\f{1}{d_1 d_2} \| N^k - N^l \|_F^2 > \f{ \ga^2 \alpha^2}{8}.
\eeq
\end{enumerate}
\end{lemma}

Then we construct the packing set $\mathcal{M}$ by setting
\be
	\mathcal{M}  = \Big\{  N + \alpha (1-\ga/2)  E_{d_1,d_2}  :  N \in \mathcal{S}(\alpha,\ga) \Big\},   \lbl{ps}
\ee
where $E_{d_1,d_2}\in \br^{d_1 \times d_2}$ is such that the $(d_1,d_2)^{th}$ entry equals one and others are zero. Clearly, $|\mathcal{M}|= | \mathcal{S}(\alpha, \ga)|$. Moreover, for any $M \in \mathcal{M}$, $M_{k,l}\in \{\alpha, (1-\ga)\alpha \}$ by the construction of $\mathcal{S}(\alpha,\gamma)$ and \eq{ps}, and
$$
	\| M \|_{\max} = \| N +\alpha(1-\ga/2) E_{d_1,d_2} \|_{\max}  \leq \f{\alpha\s{r}}{2} + \alpha(1-\ga/2) \leq \alpha \s{r},
$$
provided that $r\geq 4$. Therefore, $\mathcal{M}$ is indeed a $\de$-packing of $K_{\max}(\alpha, R)$ in the Frobenius metric with $$\de^2 = \f{\alpha^2 \ga^2 d_1 d_2}{8},$$ i.e. for any two distinct $M, M' \in \mathcal{M}$, we have $\| M - M' \|_F \geq \de$.

Next, a standard argument (e.g. \cite{YB99, Yu96}) yields a lower bound on the $\| \cdot \|_F$-risk in terms of the error in a multi-way hypothesis testing problem. More concretely,
\beq
	\inf_{\hat{M}} \max_{M \in K_{\max}(\alpha , R)} \e \| \hat{M} - M \|_F^2 \geq \f{\de^2}{4} \min_{\tilde{M}} \p(\tilde{M} \neq M^{\star}),
\eeq
where the random variable $M^{\star} \in \br^{d_1 \times d_2}$ is uniformly distributed over the packing set $\mathcal{M}$, and the minimum is carried out over all estimators $\tilde{M}$ taking values in $\mathcal{M}$. Applying Fano's inequality \cite{fano} gives the lower bound
\be
	\p(\tilde{M} \neq M^{\star})  \geq 1- \f{I(M^{\star}; Y_S) +  \log 2}{\log |\mathcal{M}|},  \lbl{fi}
\ee
where $I(M^{\star}; Y_S)$ denotes the mutual information between the random parameter $M^{\star}$ in $\mathcal{M}$ and the observation matrix $Y_S$. Following the proof of Theorem 3 in \cite{DP}, we could bound $I(M^{\star}; Y_S)$ as follows:
\beq
	\lefteqn{ I(M^{\star}; Y_S) \leq \max_{M, M' \in \mathcal{M}, M \neq M'} \mathbb{K}(Y_S|M \| Y_S | M' ) }  \\
	&= &    \max_{M, M' \in \mathcal{M}, M \neq M'} \sum_{(k,l) \in S} \mathbb{K}(Y_{k,l}| M_{k,l} \|   Y_{k,l}| M'_{k,l})   \\
	&\leq &   \f{  n [F(\alpha)-F((1-\ga)\alpha)]^2}{F((1-\ga)\alpha)[1-F((1-\ga)\alpha)]} \leq  \f{n \alpha^2 \ga^2}{\beta_{(1-\ga)\alpha}},
\eeq
where the last inequality holds provided that $F'(x)$ is decreasing on $(0, \infty)$. Substituting this into the Fano's inequality \eq{fi} yields
\beq
	\p(\tilde{M} \neq M^{\star})  \geq 1- \f{  \f{n \alpha^2 \ga^2}{ \beta_{(1-\ga)\alpha}}+  \log 2}{\f{r (d_1 \vee d_2)}{16 \ga^2}}
\eeq
Recall that $\ga^*>0$ solves the equation \eq{ga*}, i.e.
$$
	\ga^*  =  \min\bigg\{ \f{1}{2}, \,  \f{R^{1/2}}{\alpha}\bigg(\f{\beta_{(1-\ga^*)\alpha} (d_1 \vee d_2)}{32   n}  \bigg)^{1/4}  \, \bigg\}.
$$
Requiring $ \f{64 \log (2) (\ga^*)^2}{d_1 \vee d_2} \leq r \leq (d_1 \wedge d_2) (\ga^*)^2$, which is guaranteed by \eq{r}, to ensure that this probability is least $1/4$. Consequently, we have
\beq
	\inf_{\hat{M}} \max_{M \in K_{\max}(\alpha , R)} \e \| \hat{M} - M \|_F^2 \geq  \f{\alpha^2 (\ga^*)^2 d_1 d_2 }{128},
\eeq
which in turn implies \eq{minimax-lbd}.  \bbox \\

\subsection{Proof of Theorem \ref{wei}}

The proof of Theorem \ref{wei} modifies the proof of Theorem \ref{thm1-1b}, therefore we only outline the key steps in the following. Let $\{A_1, ..., A_n\}=\{(i_1, j_1), ..., (i_n, j_n)\}$ be independent random variables taking values in $[d_1] \times [d_2]$ according to $\Pi$, and recall that
\beq
	\ell_S(M;Y) = \sum_{t=1}^s  \Big[  \textbf{1}_{\{Y_{A_t}=1 \}}
	\log \Big( \f{1}{F(M_{A_t})} \Big) +
	 \textbf{1}_{\{ Y_{A_t}=-1 \}} \log \Big( \f{1}{1-F(M_{A_t})} \Big) \Big].
\eeq
According to \cite{SS2005} and the proof of Theorem \ref{thm1-1b}, it suffices to derive an upper bound on
\beq
   \De := \e \Big[ \sup_{M \in K_{\Pi, *}}
	\sum_{t=1}^n \f{\varepsilon_t}{\s{\pi_{i_t \cdot} \pi_{\cdot j_t}} }(M_{w})_{A_t}   \Big]   =    \e \Big[ \sup_{M \in K_*(\alpha, r)}
	\sum_{t=1}^n \f{\varepsilon_t}{\s{\pi_{i_t \cdot} \pi_{\cdot j_t}} }  M_{A_t}   \Big],
\eeq
where $\varepsilon_t$ are i.i.d. Rademacher random variables. Then it follows from \eq{tn.cvhull} that
\beq
	\lefteqn{ \De  \leq \alpha \s{r}  \cdot  \e \Big[ \sup_{ \|u\|_2=\|v\|_2=1}
	\sum_{t=1}^n \f{\varepsilon_t}{\s{\pi_{i_t \cdot} \pi_{\cdot j_t}} } u_{i_t}v_{j_t}   \Big] } \\
	&=&  	\alpha \s{r}   \cdot  \e \bigg[ \sup_{ \|u\|_2=\|v\|_2=1}
	\sum_{i,j} \bigg( \sum_{t: (i_t,j_t)=(i,j)} \f{\varepsilon_t}{\s{\pi_{i_t \cdot} \pi_{\cdot j_t}} } \bigg) u_iv_j   \bigg]  \\
	& = & \alpha \s{r}  \cdot  \e \Big[ \Big\|  \sum_{t=1}^n \varepsilon_t \f{e_{i_t}e_{j_t}^T}{\s{\pi_{i_t \cdot} \pi_{\cdot j_t}}} \Big\| \Big] .
\eeq
An upper bound on the above spectral norm has been derived in \cite{FSSS} using a recent result of Tropp (2012). Let $Q_t = \varepsilon_t \f{e_{i_t}e_{j_t}^T}{\s{\pi_{i_t \cdot} \pi_{\cdot j_t}}} \in \br^{d_1 \times d_2}$ be i.i.d. random matrices with zero-mean, then the problem reduces to estimate $\e \| \sum_{t=1}^s Q_t \|$. Following  \cite{FSSS}, we see that
\beq
	\e \Big\| \sum_{t=1}^n Q_t \Big\| \leq C \Big( \sigma_1 \s{\log (d)} + \sigma_2 \log (d) \Big)
\eeq
with (under condition \eq{ass2})
\beq
	\sigma_1 & =  & n \cdot \max\bigg\{ \max_k \sum_l \f{\pi_{kl}}{\pi_{k \cdot } \pi_{\cdot l}},  \,
	 \max_l \sum_k \f{\pi_{kl}}{\pi_{k \cdot } \pi_{\cdot l}} \bigg\}  \leq \mu n \max\{d_1, d_2\},   \\
	 \sigma_2 & = &  \max_{k,l}\f{1}{\s{\pi_{k \cdot } \pi_{\cdot l}}}  \leq \mu \s{d_1 d_2}  .
\eeq
Putting pieces together, we conclude that
\beq
	\De  \leq C  \alpha \s{r} \Big(  \s{\mu n \max\{d_1, d_2\}\log (d)}  + \mu \s{d_1 d_2} \log(d) \Big),
\eeq
which in turn yields that for any $\de \in (0,1)$, inequality
\beq
	\lefteqn{  \mathbb{K}_{\Pi}(F(M^*) \| F(\hat{M}_{w,tr}) ) } \\
	& \leq & C \bigg\{   L_{\alpha} \alpha \s{\f{\mu  r \max\{d_1, d_2\}\log(d)}{n}}
	+  U_{\alpha}\s{\f{\log(4/\de)}{n}}  \bigg\}
\eeq
holds with probability at least $1-\de$, provided that $n \geq \mu \min\{ d_1 , d_2\}\log(d) $.  \bbox \\

\subsection{An extension to sampling without replacement}
\label{Note}

In this paper, we have focused on sampling with replacement. We shall show here that in the uniform sampling setting,  the results obtained in this paper continue to hold if the (binary) entries are sampled without replacement. Recall that in the proof of Theorem \ref{thm1-1b}, we let $A_1, ..., A_n$ be random variables taking values in $[d_1]\times [d_2]$, $S=\{A_1, ..., A_n\}$ and assume the $A_t$'s are distributed uniformly and independently, i.e. $S \sim \Pi=\{\pi_{kl}\}$ with $\pi_{kl} \equiv \f{1}{d_1 d_2}$. The purpose now is to prove that the arguments remain valid when the $A_t$'s are selected without replacement, denoted by $S \sim \Pi_{0}$. In this notation, we have
$$
	\mathcal{D}_S = \f{1}{n} \sum_{(i,j) \in S} g(M_{i,j}; Y_{i,j})  \ \ \mbox{ and } \ \
	\mathcal{D}_{\Pi_0} = \e_{S \sim \Pi_0} [\mathcal{D }_S ] = \f{1}{d_1 d_2} \sum_{(k,l)} g(M_{k,l}; Y_{k,l}).
$$
By Lemma 3 in \cite{FS2011} and \eq{riskbd1}, for any $\de>0$,
\beq
	\sup_{M \in K_{\max}(\alpha , R)} \big(  \e_Y \mathcal{D}_{\Pi_0}(M;Y) - \e_Y \mathcal{D}_S(M;Y) \big)
	 \leq 17L_{\alpha}  R\s{\f{ d  }{n}}  + U_{\alpha}\s{\f{8 (\log(4n)+ \log(2  /\de))}{n}}
\eeq
holds with probability at least $1-\de$ over choosing a training set $S$ of $2< n\leq d_1 d_2$ index pairs according to $\Pi_0$. Next, observe that the large deviation bound \eq{concentration} for the sum of independent bounded random variables is a direct consequence of Hoeffding's inequality. To see how inequality \eq{concentration} may be extended to the current case, we start with a more general problem. Let $\mathcal{C}$ be a finite set with cardinality $N$. For $1\leq n\leq N$, let $X_1, ..., X_n$ be independent random variables taking values in $\mathcal{C}$ uniformly at random, such that $(X_1, ..., X_n)$ is a $\mathcal{C}^n$-valued random vector modeling sampling with replacement from $\mathcal{C}$. On the other hand, let $(Y_1, ..., Y_n)$ be a $\mathcal{C}^n$-valued random vector sampled uniformly without replacement. Assume that $X_i$ is centered and bounded, and write $S_X = \sn X_i$, $S_Y = \sn Y_i$. Then a large deviation bound holds for $S_X$ by Hoeffding's inequality. In the proof, the tail probability is bounded from above in terms of the moment-generating function, i.e. $m_X(\la) = \e \exp(\la S_X)$. According to the notion of negative association \cite{JP83}, it is well-known that $m_Y(\la) = \e \exp(\la S_Y) \leq m_X(\la)$, which in turn gives a similar large deviation bound for $S_Y$. Therefore, inequalities \eq{concentration} and \eq{up-2} are still valid if $\Pi$ is replaced by $\Pi_0$. Keep all other arguments the same, we then get the desired result.

\section*{Acknowledgements}

We would like to thank Yaniv Plan for helpful discussions and for pointing out the importance of allowing non-uniform sampling. A part of this work was done when the second author were visiting the Wharton Statistics Department of the University of Pennsylvania. He wishes to thank the institution and particularly the first author for their hospitality.

\end{document}